\documentclass[letterpaper, 10 pt, conference]{ieeeconf}  

\IEEEoverridecommandlockouts                              

\usepackage{graphics} 
\usepackage{epsfig} 
\usepackage{tabularx}
\usepackage{array,multirow,graphicx}
\usepackage{bm}

\usepackage[dvipsnames]{xcolor}

\renewcommand{\paragraph}[1]{\vspace{.1em}\noindent\textbf{#1}.}

\title{\LARGE \bf
Directly 3D Printed, Pneumatically Actuated Multi-Material\\ Robotic Hand
}

\begin{document}

\author{Hanna Matusik$^{1}$, Chao Liu$^{1}$, Daniela Rus$^{1}$
\thanks{$^{1}$Authors are with the Computer Science and Artificial Intelligence Laboratory (CSAIL), Massachusetts Institute of Technology, 32 Vassar St, Cambridge, MA 02139, USA {\tt\small hania@csail.mit.edu}} 
}%

\maketitle

\begin{abstract}
Soft robotic manipulators with many degrees of freedom can carry out complex tasks safely around humans.  
However, manufacturing of soft robotic hands with several degrees of freedom requires a complex multi-step manual process, which significantly increases their cost. We present a design of a multi-material 15 DoF robotic hand with five fingers including an opposable thumb. Our design has 15 pneumatic actuators based on a series of hollow chambers that are driven by an external pressure system. The thumb utilizes rigid joints and the palm features internal rigid structure and soft skin. The design can be directly 3D printed using a multi-material additive manufacturing process without any assembly process and therefore our hand can be manufactured for less than 300 dollars. We test the hand in conjunction with a low-cost vision-based teleoperation system on different tasks.   
\end{abstract}


\section{INTRODUCTION}

Soft robotic manipulators offer many advantages over standard rigid robotic hands. Soft robots are safer to use alongside humans. Soft robotic designs also are more similar to designs found in nature. However, manufacturing soft robotic manipulators is a very tedious process. In most cases, the molds need to be manufactured first. Then the manipulator components are molded using silicone and then are assembled together. Combining soft and rigid components is often difficult. Furthermore, many complex shapes cannot be molded. Also, due to this complex manufacturing process, these robots are often very expensive (e.g., over \$10,000). Additionally, many current robotic manipulators have few degrees of freedom (DoFs) and therefore lack dexterity. This prevents them from being able to perform complex tasks. 

The advances in 3D printing methods and new printing materials allow us to overcome many limitations of current soft robotic manipulators. These new methods let us directly 3D print robotic systems rather than use molding and manual assembly. This leads to huge cost and time reduction. Design iterations can be faster. Moreover, manipulators with more degrees of freedom can be easily designed and manufactured. 

In this work we propose a bio-inspired design of a soft robotic hand. Our design includes 15 DoFs, more than many current robotic manipulators. This allows our hand to be more dexterous and to be able to perform complex tasks. Our design is based on a series of hollow chambers that form a pneumatic actuator. We show how fingers and a thumb can be designed using these actuators. Then, we develop a complete robotic hand that includes rigid internal structure and joints. The whole hand is directly 3D printed using a multi-material additive manufacturing process. There is no additional assembly process required and therefore no additional human labour. The cost of manufacturing the hand is \$300, which is an order of magnitude less than other comparable robotic manipulators.

We conduct a careful evaluation of our work including the pneumatic chambers, series of pneumatic chambers, finger prototypes, opposable thumb prototypes, and finally the whole hand. We evaluate the hand on different poses and grasping tasks. We use inexpensive vision-based hand tracking available in Oculus Quest 2 headsets to control the robotic hands mounted on robotic arms. We test this system on different tasks including bi-manual tasks.

\section{RELATED WORK}



Grippers are commonly deployed in particular manipulation and grasping
scenarios for robots, including parallel jaws~\cite{velo-hand} and the
adaptive SDM hand~\cite{Dollar-sdm-hand-icra-2007} that are driven by
a single actuator. More fingers and actuators could be added to
provide more flexibility, such as the Barrett
Hand~\cite{Ulrich-upenn-barrett-hand-icra-1988}, the
iRobot-Harvard-Yale Hand~\cite{Odhner-cable-driven-hand-ijrr-2014},
and the Ocean One for marine
manipulation~\cite{Stuart-ocean-one-adaptive-hand-ijrr-2017}. These
robotic devices are aiming for a set of tasks rather than general
purpose use.

Inspired by human hands, many solutions are presented to mimic the
human hand structure to achieve human-level dexterity. It is
straightforward to drive finger joints directly through a gear or a
timing pulley. Such hands, including the DLR/HIT Hand
II~\cite{Liu-dexterous-hand-iros-2008} and
KITECH-Hand~\cite{Lee-kitchen-robotic-hand-tmech-2016}, are usually
large compared with human hands. In order to make more compact robotic
hands, cable-driven mechanisms with incorporated pulleys or springs
and the linkage-driven approach are alternatives,
e.g.,~\cite{Santina-cable-driven-hand-tro-2018,Min-cable-driven-anthropomorphic-hand-ral-2021,
  Kim-linkage-robotic-hand-nature-com-2021}. In these hands,
electronics and actuators can be installed inside a forearm to avoid
bulky robotic fingers. However, these driving mechanism are also more
sophisticated and need more maintenance.

Soft robots are inherently compliant and could be promising to provide
safer and more versatile contact with the physical
environment~\cite{rus2015design}. A universal gripper using jamming of
granular material was introduced, illuminating the potential of
amorphous structures in conforming to diverse object
shapes~\cite{Brown-granular-gripper-pnas-2010}. Inspired by the art of
origami, Li et al. crafted a "Magic-ball" gripper driven by vacuum
mechanisms with an origami inner
skeleton~\cite{Li-origami-gripper-icra-2019}. In a fusion of
compliance and rigidity, Liu et al. introduced soft robots embedding
skeletal structures, favoring cable-based controls over fluidic
actuation~\cite{Liu-modular-bio-inspired-hand-robosoft-2023}. Based on
the PneuNet technique~\cite{Ilievski-pneunet-actuator-2011}, many
pneumatically-driven soft hands have been presented. Deimel and Brock
underscored the significance of adaptability by introducing a soft
robotic hand that deftly navigates intricate grasping
scenarios~\cite{deimel2016novel}. Further amplifying the capabilities
of soft robotic grippers, Homberg et al. combined grasping with
tactile sensing and introduced proprioceptive techniques, making
robust grasping more
feasible~\cite{Homberg-soft-hand-proprioceptive-grasping-auro-2018}. Embracing
the fusion of optoelectronic sensors and soft materials, Zhao et
al. created a soft prosthetic hand that harnesses stretchable optical
waveguides~\cite{Zhao-soft-hand-optical-strain-sensing-scirobotics-2016}. More
DoFs can be added to make a more dexterous pneumatic
hand by designing separated air pipes which also requires a more
complicated fabrication
process~\cite{Zhou-pneumatic-soft-hand-ral-2018}. This work is also
based on the PneuNet design. Rather than dividing the hand into
several parts for fabrication through molding, we directly 3D print
the whole structure with multiple pneumatic actuation embedded,
including four DoFs for the thumb finger, two DoFs for all other four
fingers, and three DoFs for spreading fingers apart.


\section{ROBOTIC HAND DESIGN}

\subsection{Overview}
First, we give an overview of the single pneumatic bellow, the basis of our pneumatic actuator. Then, we discuss how we combine a series of these chambers to build an actuator. We show finger design using two of these actuators. Then, we show an opposable thumb that uses a combination of pneumatic actuators and rigid joints. Finally, we show the design of the complete hand that includes four fingers, a thumb, and a palm.

\subsection{Pneumatic Actuator}

\begin{figure}[h]
    \centering
    \includegraphics[width=0.12\textwidth]{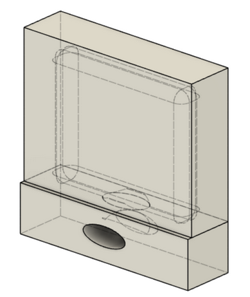}
    \includegraphics[width=0.3\textwidth]{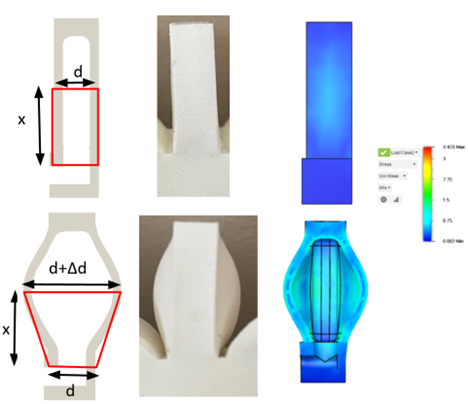}
    \caption{3D model of the bellow (first column). Cross section of the depressurized and pressurized bellow (second column). Image of the depressurized and pressurized bellow (third column). FEA analysis of the depressurized and pressurized bellow (fourth column).}
    \label{fig:bellow}
\end{figure}

There are different actuation options for inducing the movement of robotic hands (tendons, hydraulic, pneumatic, direct motors at joints, linkages). We decided to use pneumatic actuators because they can be compact and integrated into the hand with an external pressure control system. The basic primitive that we use is a pneumatic bellow designed in CAD (Autodesk Fusion 360), Figure~\ref{fig:bellow}. The bellow is made of an elastomeric material. When the bellow is in the depressurized state its width is $d$ and when it is pressurized its midsection expands by $\Delta d$. 

We simulated the bellow using finite element analysis (using a simulation in Autodesk Fusion 360). Finite Element Analysis (FEA) computes deformation of the shape given the forces acting on the interior surface of the bellow. Force is derived from pressure in the bellow. We set the internal pressure to 500 millibars and the elastic modulus to that of the material. We verified the simulation by manufacturing a bellow made of an elastomeric material and pressurizing it. The results of the experiment match the simulation well (Figure~\ref{fig:bellow}).

\begin{figure}[h]
    \centering
    \includegraphics[width=0.4\textwidth]{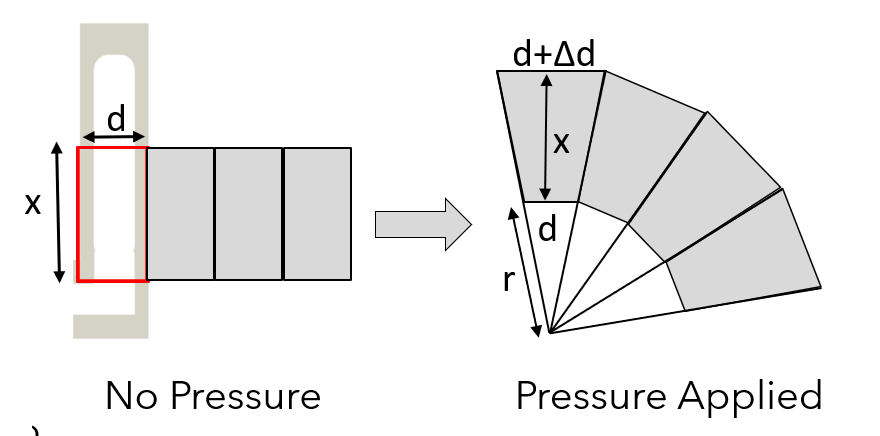}
    \caption{ Series of unpressurized bellows (left). Series of pressurized bellows(right).}
    \label{fig:bellowseries}
\end{figure}

An actuator that has 1 degree of freedom is made from a series of elastic chambers. Pressurizing the elastic chambers induces their expansion. Since the bottom of the design does not expand, the expansion of the chambers induces bending of the actuator. We derived an equation that computes the radius $r$ of the internal circle created by the bent actuator and angle $\theta$, the angle that each pressurized bellow makes the actuator bend (Figure \ref{fig:bellowseries}). These are computed using: $x$, the height from the base of the bellow to the midsection (e.g., half of the bellow height), $d$, the bellow thickness, $\Delta d$, bellow expansion:

\begin{equation}
    r = \frac{d \sqrt{x^2 + \frac{\Delta d^2}{4}}}{\Delta d}
\end{equation}
    
\begin{equation}
    \theta = 2 sin^{-1} \frac{\Delta d}{2 \sqrt{x^2 + \frac{\Delta d^2}{4}}}
\end{equation}

\begin{figure}[h]
    \centering
    \includegraphics[width=0.3\textwidth]{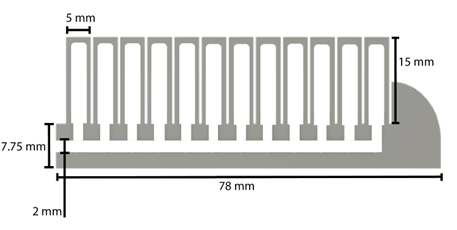}
    \caption{2D cross section of a 1 DoF pneumatic bending actuator that uses a series of chambers.}
    \label{fig:actuator}
\end{figure}

Radius, $r$, shows what the minimum diameter of an object that the hand can grasp is. Angle $\theta$ tells how much each bellow allows the actuator to bend. For example, if $x$=7.5mm, $d$=5mm, $\Delta d$=2mm, then $r$=18.9mm $\theta$= 15.2$^\circ$. This means we need 12 chambers to bend the actuator 180$^\circ$ (Figure~\ref{fig:actuator}).

There are two key parameters of this design. The first parameter is chamber wall thickness. If the wall is too thin, then the chamber can rupture. If the wall is too thick then the chamber might not deform at all. In our design we set it to 1mm - 2mm. The second key parameter is the number of chambers and the thickness of each chamber. Given a fixed length of the actuator, a thinner chamber can lead to rupture of the top surface. However, more chambers lead to more bending of the whole actuator according to the previously derived equation. Here, we set the chamber thickness to 5mm and the total number of chambers to 12-15 depending on the overall length of the actuator.

\begin{figure}[h]
    \centering
    \includegraphics[width=0.3\textwidth]{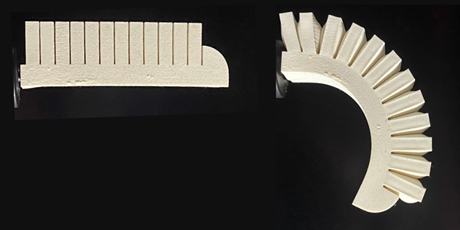}
    \caption{1 DoF pneumatic bending actuator in a depressurized and pressurized state.}
    \label{fig:actuator-bending}
\end{figure}

Typically, pneumatic actuators are manufactured by molding. First, a two-part, inverse mold is manufactured (e.g., using 3D printing) and then room temperature cured silicone or rubber is poured into the molds. The material properties of silicone are excellent (e.g., long durability, 800\% elongation at break) however, this process is usually time consuming and the produced geometries are limited. In this project, we investigated 3D printing methods to print the pneumatic actuators directly. We tested 3D printing using SLA 3D printing (Formlabs Elastic 50A resin). The material proved to be too viscoelastic and not elastic enough. We also tested printing using Stratasys Objet Connex and the Tango Plus materials. The material was too viscoelastic, and the support material could not be removed from the cavities. We settled on printing with a similar method (Inkbit, Vulcan soft elastomer).  The material properties are worse than silicone (e.g., 200\% elongation at break) but the material is soft (e.g., elastic modulus 0.53MPa) and elastomeric with good rebound (e.g., not viscoelastic). Here, we show the 3D printed actuator in the not pressurized and pressurized states (Figure~\ref{fig:actuator-bending}). Since the material is very elastic, the actuator quickly returns to its original shape when it is deflated (e.g., less than 0.3 second).

\begin{figure}[h]
    \centering
    \includegraphics[width=0.45\textwidth]{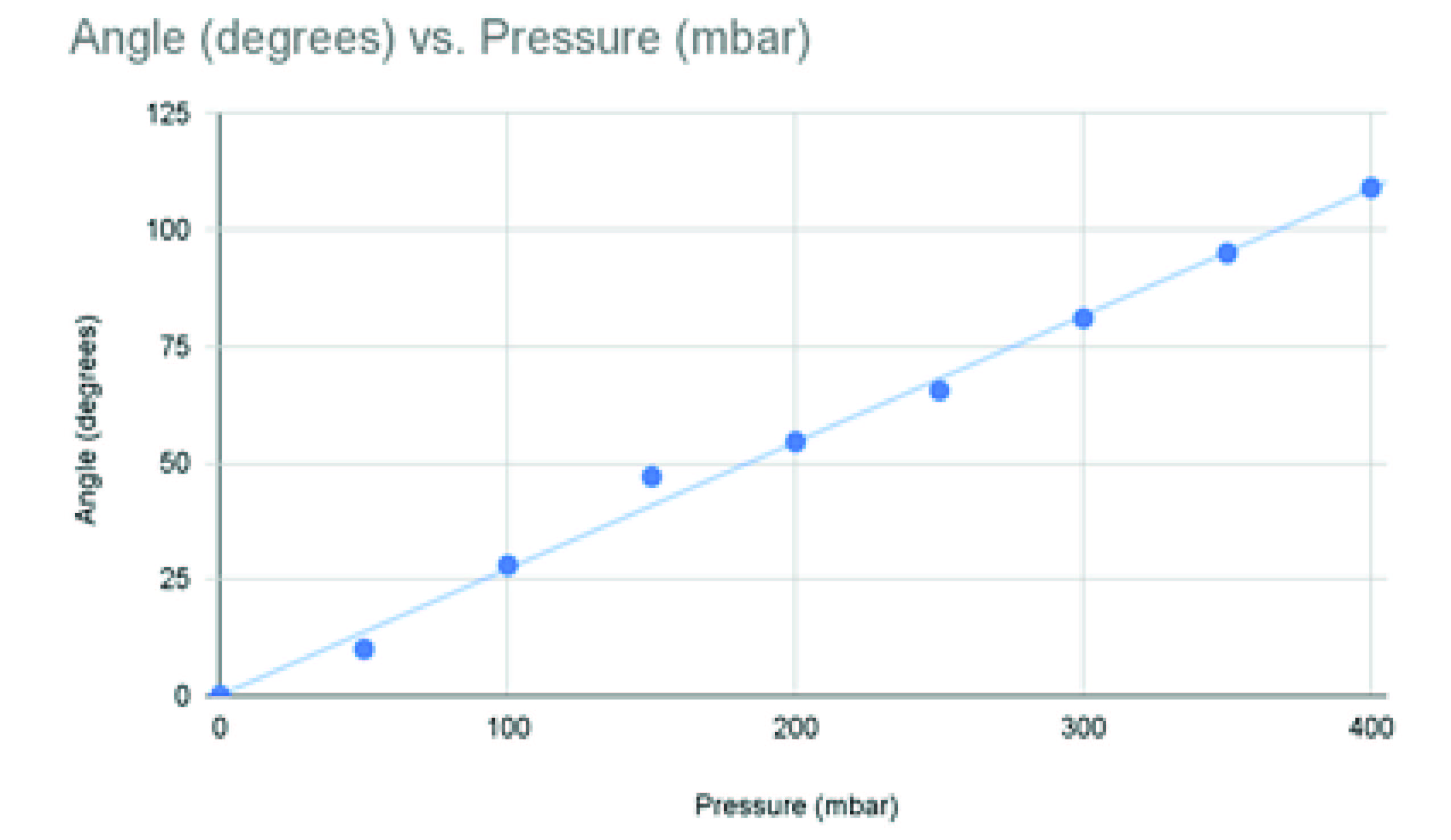}
    \caption{Relationship between internal pressure and the bending angle of the actuator.}
    \label{fig:pressure-angle}
\end{figure}

\begin{figure}[h]
    \centering
    \includegraphics[width=0.4\textwidth]{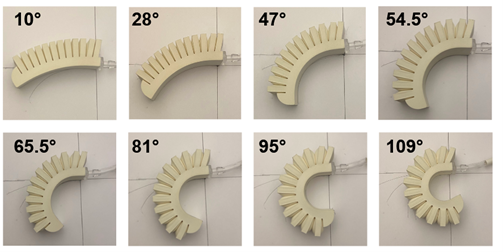}
    \caption{Images of the actuator with different internal pressure and bending angles.}
    \label{fig:pressure-angle-pic}
\end{figure}

We also capture data that translates the applied pressure to the corresponding bending angle of the actuator. The corresponding graph describes this relationship. We use a linear fit to approximate the relationship between pressure in millibars and the angle in degrees ($\theta$ = 0.272 $P^\circ$ / millibars, where $P$ is pressure in the chamber in millibars). See Figure~\ref{fig:pressure-angle} and ~\ref{fig:pressure-angle-pic}.

We test the fatigue of the actuator. In order to do this, we automatically inflate the actuator at a rate of 20 times per minute to the pressure of 400 millibars. We tested the actuator for 2 hours performing 2400 cycles total. The actuator continued to work reliably at the end of the test.

\subsection{Finger}

\begin{figure}[h]
    \centering
    \includegraphics[width=0.48\textwidth]{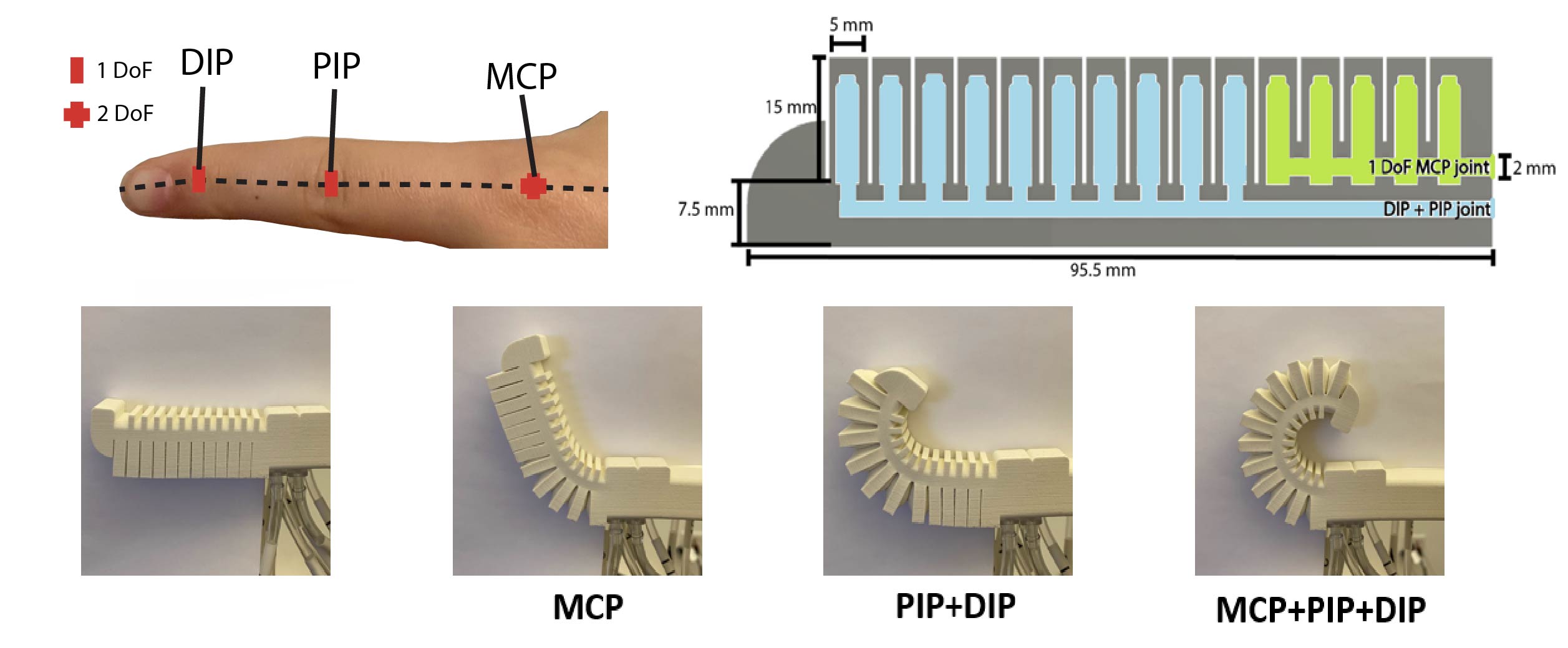}
    \caption{Left: Degrees of freedom in a human hand. Right: Cross section of a finger design with 2 DoFs: 1 DoF for MCP joint (green), 1 DoF for DIP and PIP joints (blue).}
    \label{fig:finger}
\end{figure}

Next, we moved on the design of a complete finger based on the working 1 DoF pneumatic actuator. Human fingers have three joints: the distal interphalangeal (DIP), the proximal interphalangeal (PIP), and the metacarpophalangeal (MCP). Both the DIP joint and PIP joint have 1 DoF. Typically, these two joints move together. In order to simplify the control mechanism, we decide to combine these joints together into a single pneumatic actuator. The second pressure line controls one of the DoFs of the MCP joint (Figure~\ref{fig:finger}). 

\begin{figure}[h]
    \centering
    \includegraphics[width=0.48\textwidth]{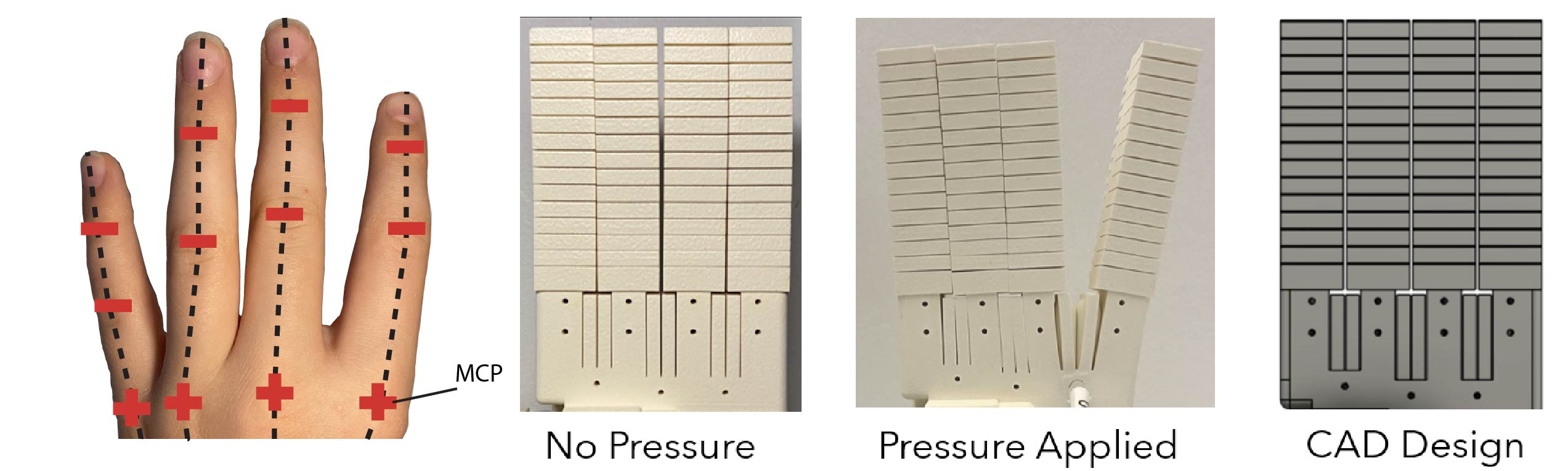}
    \caption{Left to Right: Joint in four main fingers. Manufactured four fingers. Demonstration of pressurized actuator. CAD model.}
    \label{fig:fourfingers}
\end{figure}

After completing the design of a finger, we proceed to the design of the four main fingers of the hand. We incorporate three additional actuators that correspond to the 2nd DoF of the MCP joint. Each actuator is constructed of two pneumatic chambers. These degrees of freedom allow the fingers to spread apart. See Figure~\ref{fig:fourfingers}.

\subsection{Thumb}

\begin{figure}[h]
    \centering
    \includegraphics[width=0.48\textwidth]{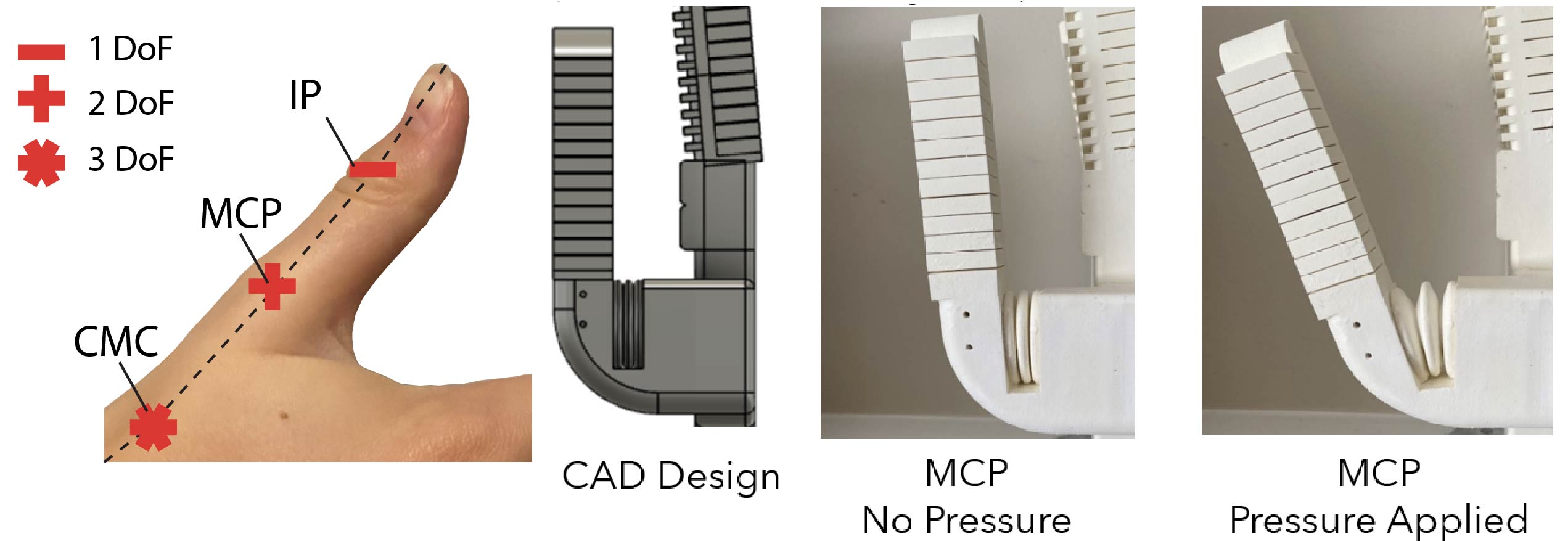}
    \caption{Design of MCP joint. Image of the joint with no pressure. Image of the joint when pressure is applied. }
    \label{fig:thumb-mcp}
\end{figure}

An opposable thumb is a vital component in human hands because it enables stronger and more stable grasps. Usually, the thumb is the most complex component of robotic hands. This is due to its high number of degrees of freedom. A thumb has 7 DoFs and three joints: two joints with three DoFs (carpometacarpal (CMC) and MCP) and one joint with one DoF (interphalangeal (IP) joint). Our simplified design is a compromise. It has only four DoFs to provide most of the functionality of a real thumb (one joint with two DoFs and two joints with one DoF each).  For the first DoF of the MCP joint and the DoF of the IP joint, we use the same design as the rest of the fingers. For the second DoF of the MCP joint, which spreads the thumb apart from the hand, we use the same design used to spread the four fingers apart. This mechanism is shown in Figure~\ref{fig:thumb-mcp}. 

\begin{figure}[h]
    \centering
    \includegraphics[width=0.5\textwidth]{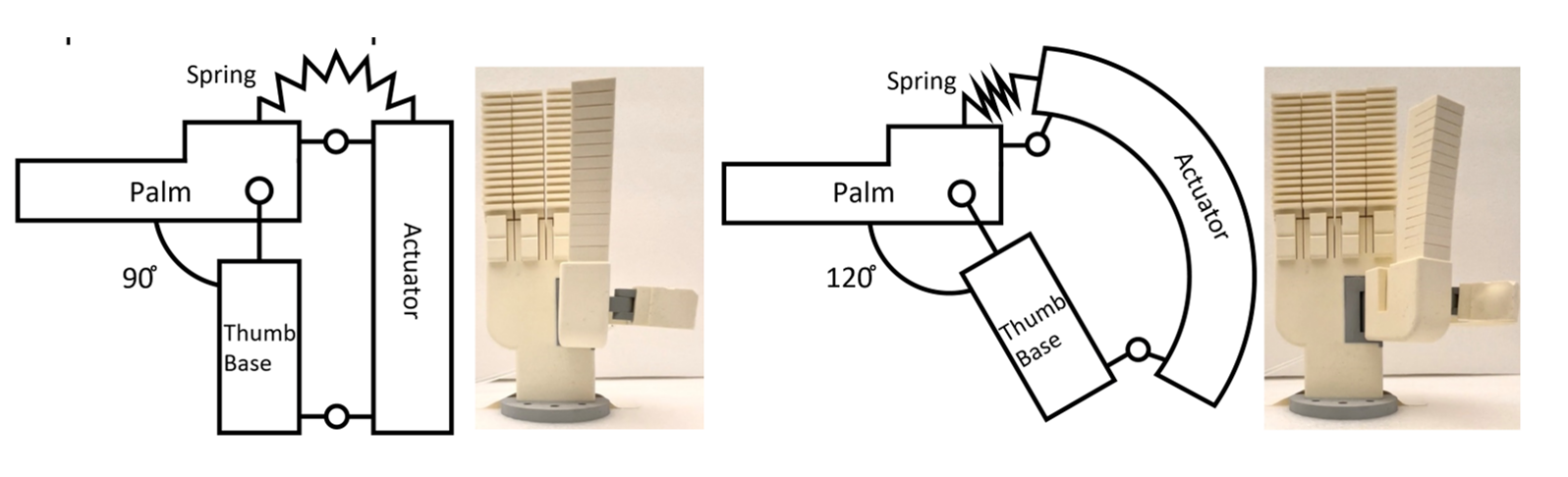}
    \caption{CMC joint mechanism. Left: Diagram and corresponding thumb position in manufactured hand when no pressure is applied. Right: Diagram and corresponding thumb position in manufactured hand when pressurized.}
    \label{fig:thumb-cmc}
\end{figure}

Finally, we add another actuator which gives functionality to one DoF of the CMC joint. When the actuator bends, the distance between the two ends decreases. In this way, the pneumatic actuator works like a human muscle, which contracts. This allows the base of the thumb to be pulled from the front towards a position where it is closer to being coplanar with the rest of the fingers. To bring the thumb back to its original position after the pressure is released, we add a soft spring made from an elastomeric material. This mechanism is shown in Figure~\ref{fig:thumb-cmc}.

\subsection{Palm}

\begin{figure}[h]
    \centering
    \includegraphics[width=0.25\textwidth]{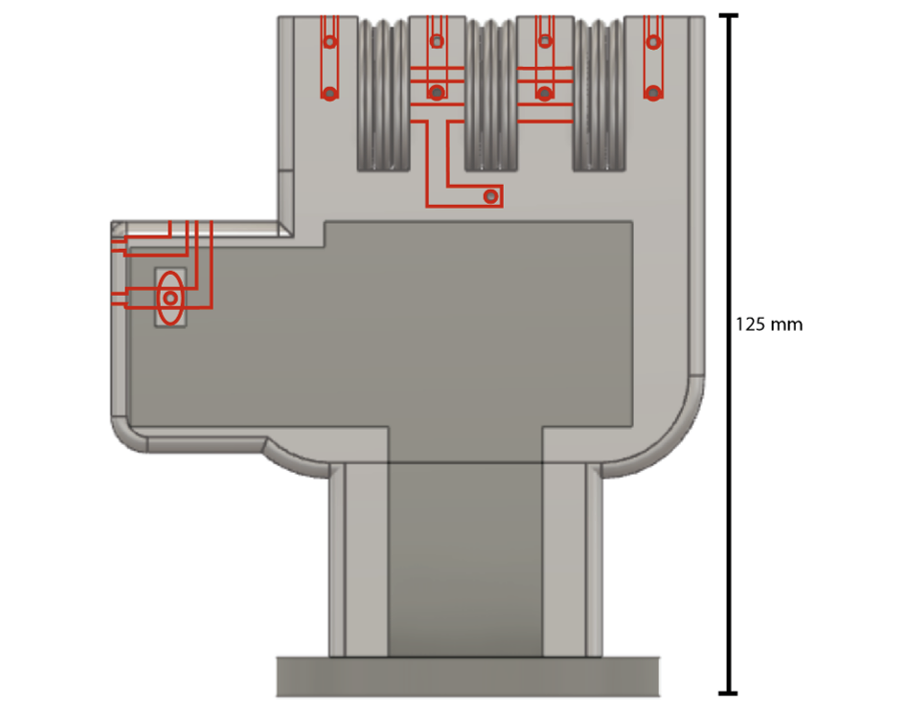}
    \caption{Design of the multi-material palm. The inner core is made of a rigid material. The skin is made of an elastomeric material.}
    \label{fig:palm}
\end{figure}

The palm is designed to mimic the shape of a human palm (Figure~\ref{fig:palm}). The rigid material is used for the inner core of the hand to provide structural strength. Without using the rigid material, the hand is too soft and could not be used to grab any heavier objects. The rigid material is also used to manufacture a stiff rigid plate that is used to attach the robot hand to an external robot arm. The palm also contains a routing network for pneumatic channels. In our design, the channel diameter is set to 3 mm to allow good air flow and reliable manufacturing. This solution simplifies connecting the hand to the external pressure control system.

\subsection{Complete Hand}

\begin{figure}[h]
    \centering
    \includegraphics[width=0.4\textwidth]{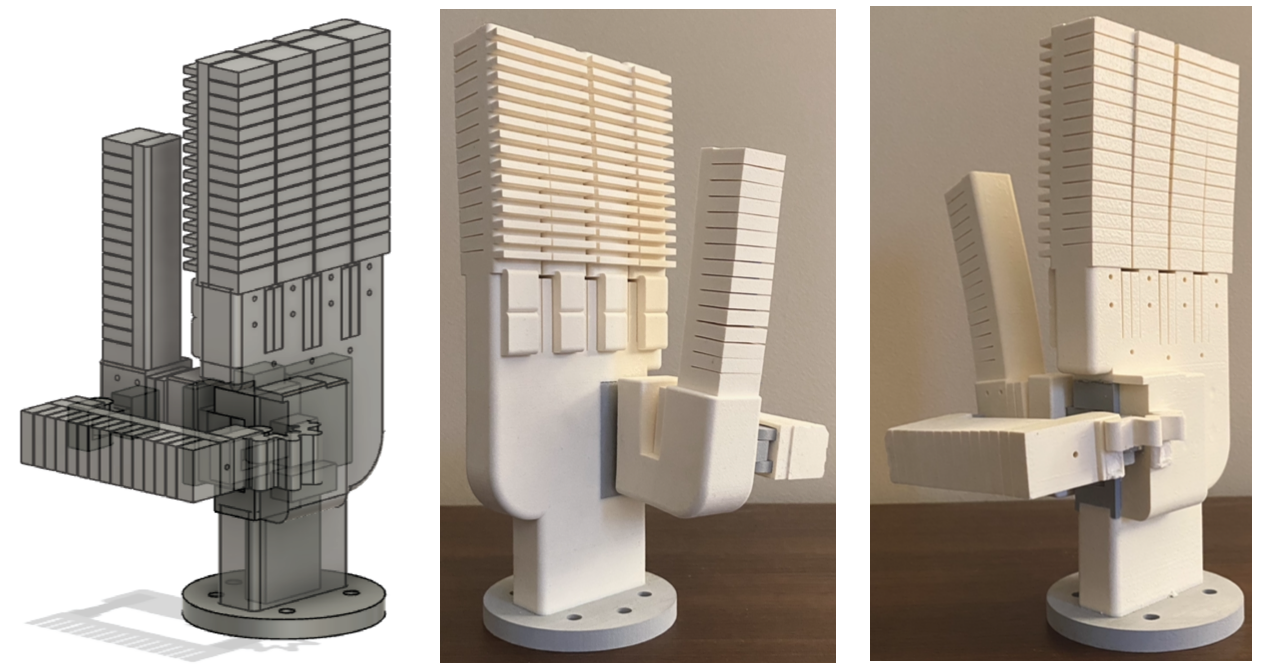}
    \caption{Whole hand: CAD model (left), 3D printed hand (right)}
    \label{fig:whole-hand}
\end{figure}

The complete hand design has 15 degrees of freedom that allow for highly dexterous movements (Figure~\ref{fig:whole-hand}). Each of the four fingers has 2 degrees of freedom (8 degrees of freedom total). There are 3 degrees of freedom that spread the fingers apart. The thumb has 4 degrees of freedom. The pneumatic actuators and the skin are made of an elastomeric material. The core of the hand and joints are made of a rigid material that provides structural integrity. This hand design can be directly manufactured using 3D printing without any additional assembly process. The overall cost is \$300. 

We connect the pressure channels of the pneumatic hand to fifteen of a 16-channel proportional valve terminal (MPA-FB-VI, Festo). The channels of the valve terminal can be individually addressed. They can supply pressures between 0 kPa to 250 kPa at a flow rate per channel of up to 380 L/min.

\begin{figure}[h]
    \centering
    \includegraphics[width=0.3\textwidth]{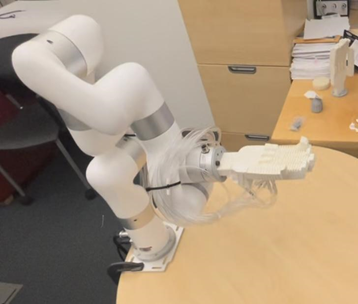}
    \caption{3D printed hand attached to the robotic arm. }
    \label{fig:arm}
\end{figure}

In order to perform complex tasks we connect the soft hand to a standard robot arm (XArm 6 by UFactory). We attach the robot hand using the mounting plate for the end-effector. We also attach the pressure lines to the robot hand. In order to perform bi-manual tasks, we use two robotic arms with mounted hands (see Figure~\ref{fig:arm}). 

\section{EVALUATION}

\subsection{Design Approach and Iterations}

We performed design iterations at different levels. At the component level, we worked on different versions of the pneumatic chamber design and the actuator design. Then, at the assembly level, we worked on different versions of the design of the whole finger, thumb, and palm. Finally, at the system level we made sure that the whole hand worked as intended. In our approach, we tested each component before moving to the next level. Design and prototyping iterations were very quick due to the use of 3D printing. Here we describe a few imported changes that we made to the design to improve its functionality. 1) The thumb originally was at the same orientation as the rest of the fingers. In the next iteration, we moved the thumb forward and turned it 90 degrees. In the next iteration, we added the rigid joint and the ability for the thumb to move between opposable and non-opposable position. 2) Originally, the fingers did not bend all the way and the hand was unable to grasp thinner objects. In the next iteration we decreased the radius by changing actuator parameters. 3) Initially, the fingers also bent very slowly. This was because the 3d printing support material (wax) from 3D printing was not completely removed. In the next iteration we made the routing larger which allowed for better rinsing of the wax, resolving the problem.

\subsection{Hand Evaluation}

\begin{figure}[h]
    \centering
    \includegraphics[width=0.42\textwidth]{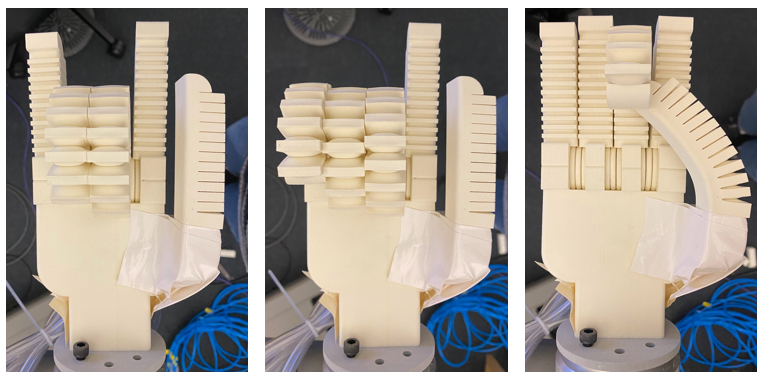}
    \includegraphics[width=0.42\textwidth]{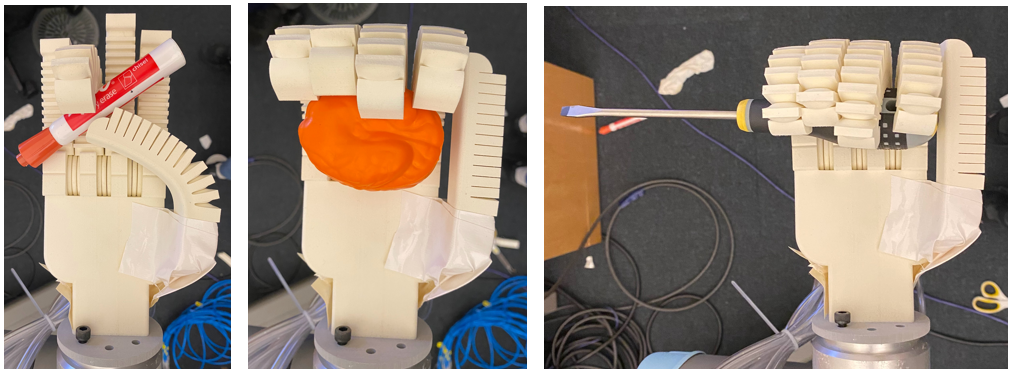}

    \caption{Evaluation: different poses (top) and grasping objects (bottom).}
    \label{fig:poses-grasps}
\end{figure}

We evaluate the operation of the hand by two tests (Figure~\ref{fig:poses-grasps}). First, we demonstrate that the hand can reach different poses like the human hand. We also demonstrate that the opposable thumb can successfully touch the tip of the other fingers. Next, we perform the evaluation by demonstrating different types of grasps. We demonstrate that the hand can hold both large objects like water bottles, small tools like screw-drivers, and larger round objects. We also demonstrate that objects like pens or markers can be held well.

\subsection{Hand Tracking and Teleoperation}

Tracking of human hands usually requires expensive hardware. A few examples of possible approaches include magnetic trackers, optical trackers, or custom gloves. Recently, new and much more inexpensive VR headsets have been developed which provide reliable and accurate hand tracking. We choose the most popular VR headset, Oculus Quest 2 by Meta. It provides realistic and accurate tracking of both human hands at more than 30 Hz. User's joint angles are estimated with less than a few degree error and hand position within a few millimeters. 

The teleoperation system is implemented in the Unity environment. The hand tracking is carried out using the Oculus SDK. Oculus SDK provides the position and orientation of the tracked human hand in the Unity coordinate system. To carry out teleoperation, we translate this position and orientation to the position and orientation of the robot end point in the robot coordinate system. We have a simple procedure to align the two coordinate systems. We position the robot hand in a start position and orientation. At the same time, we also position the human hand in a corresponding start position and orientation. The relative orientation and positions difference between two coordinate systems are stored.

In run-time, we compute the appropriate transformations to compute the position and orientation of the robot hand after the human hand is moved to a new position. Then, the robot positions and orientations are computed and sent to the robot arm using TCP/IP connection instructing the robot arm to set the location and orientation of its end effector to this location. The robot arm computes the inverse kinematics (IK). We tested sending these updates at 10 Hz. There is a delay in the system. It is less than 0.5 seconds. Similarly, the joint angles of the tracked hand are extracted. The angles are converted to pressures using the linear relationship we computed before. The pressure values are set to the appropriate registers using Modbus interface. This is also updated at 10 Hz or higher. There is almost no delay in the system thanks to Modbus communication.

\subsection{Teleoperation Results}

\begin{figure}[h]
    \centering
    \includegraphics[width=0.45\textwidth]{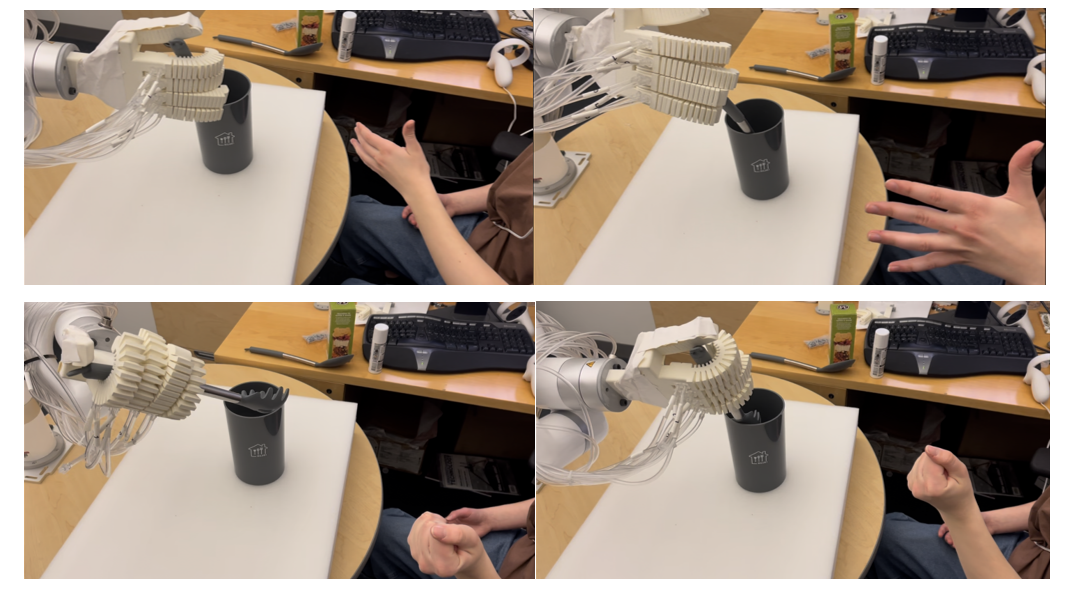}
    \caption{Teleoperation: removing a tool from a bucket and putting it back.}
    \label{fig:teleop1}
\end{figure}

\begin{figure}[h]
    \centering
    \includegraphics[width=0.45\textwidth]{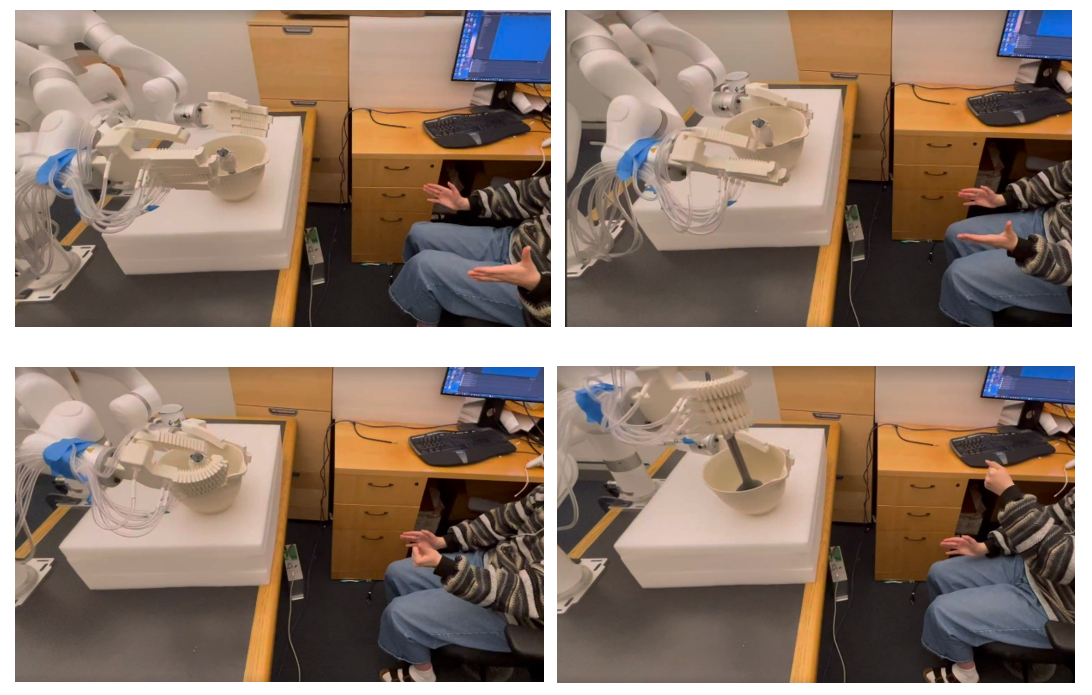}
    \caption{Bi-manual teleoperation: stirring the spoon while holding the pot. }
    \label{fig:teleop2}
\end{figure}

We experimented with teleoperation using the system (Figure~\ref{fig:teleop1}). In our setting, the robot system is positioned on one side of the table and the human sits across the table. First, we evaluated teleoperation of the movement of the hand position and orientation. As someone move his/her hand, the robot system moves the hand to the corresponding location and orientation. We also show a few tasks. For example, we show how to grab a spatula from a bin and how to put it back into the bin. We also performed an evaluation of the system using two robotic arms and hands (Figure~\ref{fig:teleop2}). We tested the system on different tasks that include sweeping the trash, moving objects (e.g., bottle with water, bowl), stirring in a pot, and dressing a mannequin. Most tasks are much easier to perform using two hands compared to using a single hand.

Our current teleoperation system works quite well. It is quite responsive and the delay between my motions and robot movements are not very noticeable. The operator does not need to wear any additional gloves that would make movements more difficult. However, it is also difficult to see whether the hand makes contact with objects. This sometimes leads to not being able to grasp objects. Tasks such as peeling a vegetable would be very challenging since human operator does not receive any haptic feedback and it is very difficult to see what the robotic hand is doing from 1.5 meters away.

\section{CONCLUSION}

In this research we developed an inexpensive directly 3D printed robotic hand with 15 degrees of freedom that can perform a large variety of household tasks.
First, we designed and fabricated an inexpensive, 3D printable, pneumatic soft actuator. Based on this actuator, we designed and fabricated pneumatically-actuated fingers with two DoFs. We have also designed a pneumatically-actuated multi-material thumb with rigid joints. Our final robotic hand has 15 degrees of freedom and costs \$300. To evaluate the hand, we developed an inexpensive teleoperation system using vision-based hand tracking. Finally, we evaluated the whole system on the teleoperation of different tasks. We believe our inexpensive robotic hand and teleoperation system is an important step for advancing and popularizing robotics at home.

There are many possible improvements of the system we have developed. The first main improvement is to increase the strength of the grasp. Currently, only light objects (less than approximately one pound) can be lifted. The current actuator needs to be redesigned to improve this limit. Second, we also plan to expand the range of tasks that are performed. We would like to test the hand on many more household tasks and determine which tasks cannot be performed. Based on the results of this evaluation, more redesign changes might be necessary. For example, tasks such as cutting bread require more force. The cutting task would be very difficult for the current hand. Tasks such as peeling a vegetable are very difficult since they require both a lot of force and fine-grained manipulation. The next major improvement is to improve the routing of pressure lines so that they can be integrated within the hand. We have also performed limited testing of the extended hand durability and material fatigue (only 2400 cycles). We plan to set up tests to carry out 10,000s of actuation cycles to determine hand durability. Based on the results of these tests some design changes might be required. The current pressure drive system is currently expensive. It would be very useful to develop a low-cost pressure control valve system. We also plan to add 3D printed sensing to the hand design. For example, pressure-based sensors can be incorporated to enable tactile sensing capabilities. Finally, the teleoperation system could also be improved by mounting a camera on the robot arm and relaying the video feed to the VR headset.


\section*{ACKNOWLEDGMENT}
This work was done with the support of National Science Foundation grant number 1830901 and the Gwangju Institute of Science and Technology.

\addtolength{\textheight}{-11cm}
\bibliographystyle{IEEEtran}
\bibliography{IEEEabrv,bibliography}

\end{document}